\pdfoutput=1
\documentclass[manuscript,nonacm]{acmart}

\usepackage{booktabs}
\usepackage{multirow}
\usepackage{array}
\newcommand{\tabitem}{~~\llap{\textbullet}~~}
\newcommand{\specialcell}[2][c]{%
    \begin{tabular}[c]{@{}#1@{}}#2\end{tabular}}%
    
\newcolumntype{C}{>{\centering\arraybackslash}m{38mm}}
\newcommand{\centercell}[1]{\multicolumn{1}{C}{#1}}

\setcopyright{none}
\acmConference{}
\acmBooktitle{}

\begin{document}

\title{Creativity and Machine Learning: A Survey}

\author{Giorgio Franceschelli}
\email{giorgio.franceschelli@unibo.it}
\affiliation{%
  \institution{Alma Mater Studiorum Università di Bologna}
  \city{Bologna}
  \country{Italy}
}

\author{Mirco Musolesi}
\email{m.musolesi@ucl.ac.uk}
\affiliation{%
  \institution{University College London}
  \city{London}
  \country{United Kingdom}
}
\affiliation{%
  \institution{Alma Mater Studiorum Università di Bologna}
  \city{Bologna}
  \country{Italy}
}

\renewcommand{\shortauthors}{Franceschelli and Musolesi}

\begin{abstract}
  There is a growing interest in the area of machine learning and creativity. This survey presents an overview of the history and the state of the art of computational creativity theories, key machine learning techniques (including generative deep learning), and corresponding automatic evaluation methods. After presenting a critical discussion of the key contributions in this area, we outline the current research challenges and emerging opportunities in this field.
\end{abstract}

\maketitle

\section{Introduction}

The connection between creativity and machines is as old as computer science itself. In 1842, Lady Lovelace, an English mathematician and writer, who is recognized by many as the first computer programmer, issued what is now known as ``Lovelace’s Objection'' \cite{turing50}. She stated that the Analytical Engine (the digital programmable machine proposed by Charles Babbage \cite{babbage64}) \textit{``has no pretensions to originate anything. It can do whatever we know how to order it to perform''} \cite{menabrea43}. Indeed, in the centuries that followed, numerous projects and studies have been undertaken with the aim of designing machines capable of ``originating something'' \cite{cohen88, cope89, hofstadter94, langley87, meehan77, racter84}. We have witnessed the emergence of a specialized field in computer science, namely Computational Creativity \cite{cardoso09}, which concerns the study of the relationship between creativity and artificial systems \cite{colton12a,wiggins06}.

In this context, the adoption of deep learning (DL) techniques has led to substantial breakthroughs in recent years. Vast computational power and very large amounts of available data are at the basis of the increasing success of deep generative models (i.e., generative models based on DL \cite{foster19}). Indeed, generative deep learning technologies have been used to write newspaper articles\footnote{\url{www.theguardian.com/commentisfree/2020/sep/08/robot-wrote-this-article-gpt-3}}, generate human faces \cite{karras21} and voices \cite{li19}, design drugs and proteins \cite{jumper21}, and even create artworks sold for hundred thousand dollars\footnote{\url{www.christies.com/features/a-collaboration-between-two-artists-one-human-one-a-machine-9332-1.aspx}}.
While it is apparent that current technologies are able to generate impressive outputs, at the same it is also possible to argue that they cannot be considered \textit{creative} in general \cite{bown21}. In fact, the goal of generative deep learning is to produce synthetic data that closely resemble real ones fed in input \cite{foster19}. On the other hand, creativity involves novelty and diversity \cite{newell62}. While for some problems mere content generation \cite{ventura16} might be sufficient, for other tasks, e.g., in the Arts, the ability to create different (but still valuable) outputs is essential: a creative model can find practical applications in the arts and industrial design as support for artists, content creators, designers and researchers, just to name a few. Moreover, generating more diverse data might mitigate legal and ethical issues related to content reproduction \cite{henderson23,weidinger22}.

\noindent \textbf{Goal and contributions of the survey.} The goal of this survey is to present and critically discuss the state of the art in generative deep learning from the point of view of machine creativity. 
Moreover, to the best of our knowledge, this is the first survey that explores how current DL models have been or can be used as a basis for both generation (i.e., producing creative artifacts) and evaluation (i.e., recognizing creativity in artifacts). 
The contribution of this survey can be summarized as follows. After a brief overview of the meaning and definitions of creativity in Section \ref{whatcreativityis},
Section \ref{generativemodels} presents an in-depth analysis of generative deep learning through the introduction of a new taxonomy and a critical analysis of machine creativity. Then, several machine learning (ML)-based methodologies for evaluating creativity are presented and discussed in Section \ref{creativitymeasures}. Finally, Section \ref{conclusion} concludes the paper, outlining open questions and research directions for the field.

\noindent \textbf{Related surveys.} We now provide an overview of other surveys in areas related to the present work. For readers interested in a survey on deep generative models, we recommend \cite{bondtaylor21,hashvardhan20}; for an analysis of the state of the art in evaluation in computational creativity, \cite{lamb18} is an essential reading; for a review on AI and creativity in general, we recommend~\cite{rowe93}; for a practical view of generative deep learning, we suggest \cite{foster19}; finally, for an in-depth examination of artistic AI works (also in human-computer co-creations \cite{guzdial19}), \cite{miller19} is a very comprehensive source of information.

\section{Defining Creativity} \label{whatcreativityis}

Creativity has been studied for decades, and yet, there is no agreement about its definition. More than one hundred definitions have been provided \cite{aleinikov00,treffinger96}, and the number is still growing. In other words, we can say that creativity is a suitcase word, i.e., people conflate multiple meanings into it \cite{minsky06}. Nonetheless, some concepts are nowadays widely accepted. One of them is the possibility of studying creativity from four different perspectives: \textit{person}, \textit{press}, \textit{process}, and \textit{product} \cite{rhodes61}. These have also been studied in computational creativity \cite{jordanous16}. However, the focus has traditionally been on the \textit{product} dimension.
Indeed, the idea is that we study creativity without considering the inner being of the creator (person), the relation with the environment (press), or if the process maps to the steps a human goes through in the act of creation (e.g., \cite{amabile83}). Even if they are important, we focus on aspects of creative work in relation to the output (product) itself, and on aspects of the generative process that may or may not lead to a creative product.

For this reason, we consider Boden's three criteria for studying machine creativity, defined as \textit{``the ability to come up with ideas or artifacts that are} new, surprising and valuable\textit{''} \cite{boden03}. In particular, value refers to utility, performance, and attractiveness \citep{maher10}; it is related with both the quality of the production, and its acceptance by the society. Novelty refers to the dissimilarity between the produced artifact and other examples in its class \citep{ritchie07}. Finally, surprise refers to the degree to which a stimulus disagrees with expectation \citep{berlyne71}. We underpin our analysis on Boden's three criteria since they have been widely adopted. Boden also suggests that three forms of creativity can be identified \cite{boden03} to describe how a novel and surprising product is obtained. The three forms of creativity are \textit{combinatorial}, \textit{exploratory} and \textit{transformational}{, ordered by increasing rarity and produced surprise}. Combinatorial creativity is about making unfamiliar combinations of familiar ideas, e.g., analogies in textual forms or collages in the visual arts. Exploratory creativity involves the exploration of the conceptual space defined by the cultural context considered, e.g., inventing a new type of cut for fries. Transformational creativity involves changing that space in a way that allows new and previously inconceivable thoughts to become possible, as it has been for free verse in poetry or abstract painting in art.

Finally, it is worth noting that Boden also identified four different questions that emerge when studying computational creativity. These are referred to as Lovelace questions, because many people would respond to them by using Lovelace's objection. The first question is whether computational ideas can help us understand how human creativity works. The second is whether computers could ever do things that at least appear to be creative. The third is whether a computer could ever appear to recognize creativity. Finally, the fourth is whether computers themselves could ever really be creative, i.e., with the originality of the products only deriving from the machine itself \cite{boden03}.
While the first one is studied in Boden's work, we will provide the reader with an overview of the techniques that can possibly be used to answer the second (Section \ref{generativemodels}) and the third (Section \ref{creativitymeasures}). With respect to the fourth, Boden states that \textit{``it is not a scientific question as the others are, but in part a philosophical worry about ``meaning'' and in part a disguised request for a moral political decision''}. We agree with this position, but we hope that our survey will provide the reader with elements for answering the fourth one as well.

\section{Generative Models} \label{generativemodels}

A generative model can be defined as follows: given a dataset of observations $X$, and assuming that $X$ has been generated according to an unknown distribution $p_{data}$, a generative model $p_{model}$ is a model able to mimic $p_{data}$. By sampling from $p_{model}$, observations that appear to have been drawn from $p_{data}$ can be generated \cite{foster19}. Generative deep learning is just the application of deep learning techniques to form $p_{model}$.

At first glance, this definition appears to be incompatible with those presented in Section \ref{whatcreativityis}. Indeed, mimicry is the opposite of novelty. However, what a generative model should aim at mimicking is the underlying distribution representative of the artifacts, and not the specific artifacts themselves; in other words, it should aim at learning the conceptual space defined by the cultural context considered. A generative model can be said to exhibit combinatorial creativity if it can sample new and valuable works that are combinations of real data and exploratory creativity if the works actually differ from real ones. Vice versa, transformational creativity emerges if and only if the distribution for sampling diverges in some way from the underlying one (e.g., due to a different training process, by altering the distribution after learning, or by changing the sampling technique). In summary, what matters is how the space of solutions is learned and how artifacts are sampled from it.

\begin{figure}[ht]
  \centering
  \includegraphics[width=1\linewidth]{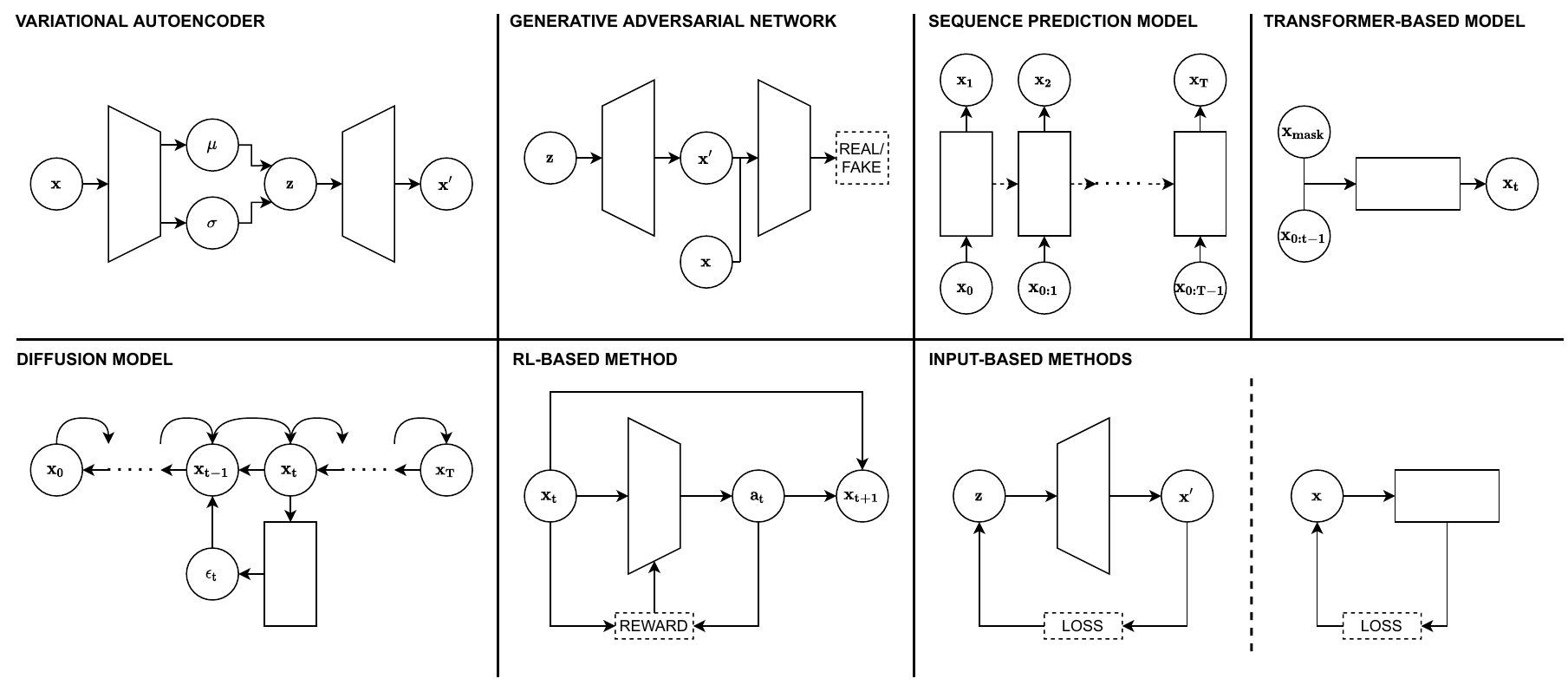}
  \caption{A schematic view of the seven classes of generative learning methods presented in this survey. Top, left to right: Variational Autoencoder (\ref{vae}), with a decoder generating $\mathbf{x'}$ given a latent vector $\mathbf{z}$, and an encoder representing $\mathbf{x}$ into a latent distribution; Generative Adversarial Network (\ref{gan}), with a generator to produce $\mathbf{x'}$, and a discriminator to distinguish between real $\mathbf{x}$ and synthetic $\mathbf{x'}$; Sequence prediction model (\ref{sequenceprediction}), with a generator to output $\mathbf{x}$ one token after the other given in input previous tokens; Transformer-based model (\ref{transformers}), with a Transformer outputting $\mathbf{x}$ one token after the other given in input previous tokens, or a masked version of $\mathbf{x}$. Bottom, left to right: Diffusion model (\ref{diffusion}), with a model to learn an error $\mathbf{\epsilon}$, which is used to incrementally reconstruct $\mathbf{x_0}$; Reinforcement Learning (RL)-based method (\ref{rlbased}), with a generative model acting (i.e., progressively generating $\mathbf{x}$) to maximize a given reward function; Input-based methods (\ref{inputbased}), with an input optimized by a given loss. The input can be a vector $\mathbf{z}$ given to a generative model to obtain the desired output, or directly a product $\mathbf{x}$ becoming the desired output.}
  \Description{Summary of Classes of Generative Learning Methods.}
  \label{fig:models}
\end{figure}

In this section, we aim at studying the level of creativity of existing generative deep learning models. Following the discussion above, we analyze how the models learn their spaces of solutions and how the observations are generated from them. A new generative deep learning taxonomy is then introduced based on the different training and sampling techniques at the basis of each method. Figure \ref{fig:models} provides a summary of the seven generative classes considered in this survey. Since our focus is on machine creativity, we do not discuss the implementation details of each class of methods. We instead present the \textit{core concepts} at the basis of each class; some relevant \textit{examples of models}; potential \textit{applications}; and a \textit{critical discussion} evaluating the level of machine creativity considering the definitions above.
As a final remark, it is worth noting that we limit our examples to the Arts (e.g., poems, music, or paintings). Indeed, generative learning can be applied to design \cite{gero00, maher12}; game content generation (see \cite{liu21} for a comprehensive survey); recipes \cite{morris12, varshney19a}; scientific discovery \cite{colton08,schmidt09}; and in general to any activity, which has a non-trivial solution \cite{brown20}.

\subsection{Variational Auto-Encoders} \label{vae}

\subsubsection{Core Concepts}

A Variational Auto-Encoder (VAE) \cite{kingma14,rezende14} is a learning architecture composed of two models: an encoder (or recognition model) and a decoder (or generative model). The former compresses high-dimensional input data into a latent space, i.e., a lower-dimensional space whose features are not directly observable, yet provide a meaningful representation. The latter decompresses the representation vector back to the original domain \cite{foster19}. Classic autoencoders directly learn to represent each input in a latent representation vector. Conversely, VAEs learn a (Gaussian) distribution over the possible values of the latent representation, i.e., the encoder learns the mean and the (log of the) variance of the distribution. 

VAEs are trained by optimizing two losses: the reconstruction loss and the regularization loss. The former is the log-likelihood of the real data $\mathbf{x}$ from the decoder given their latent vectors $\mathbf{z}$, i.e., it is the error of the decoder in reconstructing $\mathbf{x}$. The latter is the Kullback-Leibler (KL) divergence between the distribution learned by the encoder and a prior distribution, e.g., a Gaussian. Notably, the latent vector $\mathbf{z}$ in input to the decoder is obtained by means of the so-called \textit{reparameterization trick}, i.e., by sampling from the distribution defined by the mean and the variance. Without it, sampling would induce noise in the gradients required for learning \cite{kingma19}.

The mathematical derivation of the whole loss has its roots in variational inference \cite{jordan99}. Indeed, VAEs can be seen as an efficient and stochastic variational inference method, in which neural networks (NNs) and stochastic gradient descent are used to learn an approximation (i.e., the encoder) of the true posterior \cite{ganguly21}.
In VAEs, similar high-dimensional data are mapped to close distributions. This makes it possible to sample a random point $\mathbf{z}$ from the latent space, and still obtain a comprehensible reconstruction \cite{foster19}. On the other hand, VAE tends to produce blurred images \cite{zhao17}. It may also happen that high-density regions under the prior have a low density under the approximate posterior, i.e., these regions are not decoded to data-like samples \cite{aneja21}. Finally, the objective can lead to overly simplified representations without using the entire capacity, obtaining only a sub-optimal generative model \cite{burda16}.

\subsubsection{Examples of Models}

Several models based on VAEs have been proposed \cite{kingma19} in recent years. In the following, we focus on those relevant to our discussion on machine creativity.
In $\beta$-VAE \cite{higgins17}, a parameter $\beta$ is used to scale the magnitude of the regularization loss, which allows a better disentanglement of the latent space \cite{burgess18}. Another example is VAE-GAN \cite{larsen16}, which merges VAE and Generative Adversarial Networks (GAN; see Section \ref{gan}) \cite{goodfellow14}. This is done by treating the decoder as the generator of the GAN, thus training it by means of the GAN loss function. This leads to the generation of substantially less blurred images. Similarly, Adversarially Learned Inference (ALI) \cite{dumoulin17} merges VAE and GAN by asking the discriminator to distinguish between pairs of real data (and their latent representations) and pairs of sampled representations and synthetic data. Instead, Adversarial Autoencoders (AAE) \cite{makhzani16} substitute the regularization loss with a discriminative signal, where the discriminator has to distinguish between random latent samples and encoded latent vectors. Another way to address the problem of ``sample blurriness'' is with PixelVAE \cite{gulrajani17}, where the autoregressive PixelCNN \cite{oord16a,oord16b} is used as the decoder. In \cite{bowman16}, to deal with sequential data such as texts, where generation requires more steps, the encoder learns to produce a latent representation of a sentence, while the recurrent neural network RNN-based decoder learns to reproduce it word after word. However, VAE can also generate text by means of convolution and deconvolution \cite{semeniuta17}.
To solve the problem of low-density regions, the authors of \cite{aneja21} propose an energy-based model called noise contrastive prior (NCP), trained by contrasting samples from the aggregate posterior to samples from a base prior. Finally, another interesting model is Vector Quantised-VAE (VQ-VAE) \cite{oord17}; in this case, the encoder outputs discrete, rather than continuous, codes, and the prior is learned rather than static.

\subsubsection{Applications}

VAEs can be used for semi-supervised classification to provide an auxiliary objective, improving the data efficiency \cite{kingma14b,maaloe16}; to perform iterative reasoning about objects in a scene \cite{eslami16}; to model the latent dynamics of an environment \cite{watter15}.
Of course, VAEs have also been used to generate synthetic data, including for conditional generation. For example, a layered foreground-background generative model can be used to generate images based on both the latent representation and a representation of the attributes \cite{yan16}. 
In \cite{gomezbombarelli18} the latent space of a VAE is trained on chemical structures by means of gradient-based optimization toward certain properties (see Section \ref{inputbased}). AAEs have also been applied to the same problem \cite{kadurin17}. Finally, another interesting application of VAE is Deep Recurrent Attentive Writer (DRAW) \cite{gregor15}. DRAW constructs scenes in an iterative way, by accumulating changes emitted by the decoder (then given to the encoder in input). This allows for iterative self-corrections and a more natural form of image construction. RNNs and attention mechanism are used to consider previous generations and to decide at each time step where to focus attention, respectively.

\subsubsection{Critical Discussion}

Models based on VAEs can be considered as an example of exploratory creativity. The latent space is learned with the goal of representing data in the most accurate way. The random sampling performed during generation is therefore an exploration of that space: regions not seen during training can be reached as well, even though they can lead to poor generation~\cite{aneja21} and some more complex variants may be needed, as discussed. On the other hand, there is no guarantee that the results will be valuable, novel, or surprising.
There is no guarantee that the generation from random sampling is of good quality or diverse from training data. Indeed, given their characteristics, VAEs discourage novelty in a sense. In particular, diversity could be achieved in theory using VAEs and gradient-based optimization techniques, such as those presented in~\cite{gomezbombarelli18}, with novelty and surprise as target properties. We will discuss these aspects in Section \ref{creativityoriented}.

\subsection{Generative Adversarial Networks} \label{gan}

\subsubsection{Core Concepts}

A Generative Adversarial Network \cite{goodfellow14} is an architecture composed by two networks: a generative model and a discriminative model. The latter learns to distinguish between real samples and samples generated by the former. In parallel, the former learns to produce samples from random noise vectors such that they are recognized as real by the latter. This competition drives both models to improve their methods until the generated samples are indistinguishable from the original ones. 

The adversarial training allows the generator to learn to produce seemingly real samples from random noise without being exposed to data. The simplicity of the idea and the quality of results are at the basis of the success of GANs. However, few limitations exist. For instance, GAN can suffer from mode collapse, where the generator only learns to produce a small subset of the real samples \cite{metz17}. In addition, the latent space of random inputs is typically not disentangled and it is necessary to introduce constraints in order to learn an interpretable representation~\cite{kazemi19}.

\subsubsection{Examples of Models}
The number of proposed variants is still growing
. An in-depth survey on GANs is~\cite{gui21}. 
Indeed, several refinements have been proposed in the past years, such as using deep convolutional networks~\cite{radford16} or self-attention~\cite{zhang18a}, incrementally growing the networks \cite{karras18}, or scaling the model parameters \cite{brock18}. In the following, we present examples that are relevant to the issue of machine creativity.

The problem of non-meaningful representation has been addressed in different ways. For instance, InfoGAN \cite{chen16} adds a latent code $\mathbf{c}$ to $\mathbf{z}$. An auxiliary model learns to predict $\mathbf{c}$ given the sample generated by means of it. In this way, it can learn disentangled representations in a completely unsupervised manner. Another possibility is Bidirectional GAN (BiGAN) \cite{donahue17}. In order to include an inverse mapping from data to latent representation, an encoder is added to the architecture. The discriminator is then trained to distinguish between pairs of random noise and synthetic data and pairs of real data and latent encoding.
It is possible to condition the generation by means of a target content \cite{odena17}, a text \cite{reed16}, or even an image \cite{isola17}. In order to do so, it is sufficient to use the conditional information as input for both generator and discriminator~\cite{mirza14}. Similarly, image-to-image translation is possible also without paired datasets. CycleGAN \cite{zhu17} trains two generators (from one domain to another, and vice versa) so that each of them produces images both from the target domain and correctly reconstructed by the counterpart.

In StyleGAN~\cite{karras19, karras20}, the generator architecture is re-designed in order to control the image synthesis process. The style of the image is adjusted at each layer based on the latent code (the specific intermediate code to control each layer is provided by a non-linear mapping network). This allows for the automatic separation of high-level attributes from stochastic variations in the generated images. It also allows for mixing regularization, where two latent codes are used alternatively to guide the generator. StyleGAN-V \cite{skorokhodov22} builds on top of it to learn to produce videos by only using a few frames of it. To generate longer and more realistic motions, a two-stage approach can be used as well: first, a low-resolution generator is adversarially trained on long sequences; then, a high-resolution generator transforms a portion of the produced, low-resolution video in a high-resolution one \cite{brooks22}.

Finally, it is also worth mentioning variants that adapt GANs to sequential tasks (e.g., text generation). Since GANs require the generator to be differentiable, they cannot generate discrete data \cite{goodfellow17}. However, several techniques have been proposed to avoid this problem. One possibility is to transform the discrete generation into a continuous one. Music can be processed like an image by considering its waveform (as in WaveGAN \cite{donahue18} and GANSynth \cite{engel19}) or its musical score composed of tracks and bars (as in MuseGAN \cite{dong18}). Music in a desired style can be obtained through conditional inputs.
Another possibility is to consider a soft-argmax function as an approximation of the inference for each step \cite{zhang16}. TextGAN \cite{zhang17} uses it together with feature matching to learn the production of sentences. In place of the discriminative signal, it uses the difference between the latent feature distributions of real and synthetic sentences learned by the discriminator.
Another solution is to transform the GAN into a Reinforcement Learning (RL) framework, as in Sequential GAN (SeqGAN)~\cite{yu16}. The generative model is the agent; the tokens generated so far form the state; the selection of the next token to be generated is the action to be performed; and the discriminative signal is the reward. The REINFORCE algorithm \cite{williams1992} can then be used to adversarially train the generative model. Other policy gradient methods can be used as well \cite{fedus18}. 
On the other hand, the learning signal (i.e., the reward) might be very sparse. A way to solve this issue is to use inverse RL~\cite{ziebart08}. For example, the authors of~\cite{shi18} use inverse RL to learn a reward function able to associate positive rewards to real state-action pairs, and non-positive rewards to synthetic state-action pairs. Notably, this can help solve mode collapse too. Another variant is LeakGAN \cite{guo18}. Here, a hierarchical generator composed of a Manager and a Worker is used. The Worker produces a sentence conditioned by a goal vector provided by the Manager. The Worker and the discriminative model are trained following SeqGAN; the Manager is trained to predict goal vectors that lead to the identification of advantageous directions in the discriminative feature space. More specifically, the Manager receives a feature vector from the discriminator, i.e., its last convolutional layer, at each generated token. By means of this \textit{leaked} information and the hierarchical architecture, LeakGAN produces longer and higher-quality texts. Finally, another possibility is to use Gumbel-softmax relaxation \cite{jang17, maddison17}, as in Relational GAN (RelGAN) \cite{nie19}. Controlled TExt generation Relational Memory GAN (CTERM-GAN) \cite{betti20} builds on the latter by also conditioning the generator on an external embedding input. In addition, it uses both a syntactic discriminator to predict whether a sentence is correct and a semantic discriminator to infer if a sentence is coherent with the external input.

\subsubsection{Applications}

GANs have been applied to a variety of practical problems in several application scenarios. They have been widely used for semi-supervised learning \cite{odena16}; for generating adversarial examples \cite{xiao18} to better train image classifiers \cite{madry18}; and, in general, in computer vision (see \cite{wang21} for a detailed discussion). The generative power of GANs has also found its place in recommender systems (see \cite{deldjoo21}) to generate fashion items; in science and chemistry \cite{mendezlucio20,motamed21}. Of course, its ability to generate high-quality samples has been exploited in many other areas, from anime design \cite{jin17} and 3D object modeling \cite{wu16} to photo-realistic consequences of climate change \cite{schmidt22}. Conditional inputs also allow the production of artistic works by controlling stylistic properties such as genre \cite{tan17} or influencer \cite{chu18}.
Finally, the most famous example of the artistic power of GAN is the collection of paintings by Obvious, a French art collective \cite{vernier20}; one of their works has been sold to more than 400,000 dollars\footnote{Fun fact: the sold painting is called \textit{Portrait of Edmond De Belamy} because Belamy sounds like \textit{bel ami}, a sort of French translation of... \textit{Goodfellow}.}. 

\subsubsection{Critical Discussion} \label{gandiscussion}

GANs are difficult to evaluate from a machine creativity perspective. The generator does not receive the original works as input, so it samples from a conceptual space that is built only indirectly from them. In rare cases, this can also lead to a different conceptual space (with respect to the original one) and so to transformational creativity, but it typically leads to exploratory creativity. In fact, since the goal is to learn to generate seemingly real artifacts from a latent distribution, it is likely that it will approximate the real one. Still, it is possible to identify potential creative solutions among those generated by the model.

An advantage of GANs is the presence of a \textit{recognition} network, i.e., the discriminator, trained to recognize real (valuable) works. This is important for two reasons. It suffices for being able to define GANs \textit{appreciative} \cite{colton08}, which is a central sub-task of creativity \cite{amabile83,gaut03}. In addition, it allows us to consider their products as valuable, as it is in a sense their intrinsic objective. However, there is no guarantee that they will also be new and surprising.
Nevertheless, it seems possible to extend a GAN objective to include such properties as well (see Section \ref{creativityoriented} for a discussion).

\subsection{Sequence Prediction Models} \label{sequenceprediction}

\subsubsection{Core Concepts}

A sequence prediction model is a generative model that considers generation as a sequential process. It works in an autoregressive fashion: it predicts the future outcome of the sequence (i.e., the next token) from the previously observed outcomes of that sequence, usually by means of an internal state that encodes information from the past. It is trained to minimize the prediction error for each token in the dataset. At inference time, this simple yet effective approach only requires to produce one token after the other, feeding back to the model what has been produced so far \cite{karpathy15}. It makes it possible to learn dependencies between tokens in real data so that the same dependencies can be exploited when generating synthetic data. However, this causes the generation to be highly dependent on real data, e.g., there is the risk of potentially reproducing portions of the training set.

\subsubsection{Examples of Models}

Several models have been proposed, most of them based on RNN, and especially on long short-term memory (LSTM) \cite{hochreiter97}. The reason is that RNNs use internal states based on previous computation: inputs received at earlier time steps can affect the response to the current input. However, RNNs tend to perform worse with longer sequences~\cite{bengio94}. LSTM is a specific RNN architecture that addresses the problem of long-term dependencies through the use of additional gates determining what to remember and what to forget at each step.

RNNs can be used to model joint probabilities of characters (Char-RNN) \cite{karpathy15}; words \cite{potash15}; phonemes \cite{hopkins17}; syllables \cite{zugarini19}; and even tokens from transcriptions of folk music (Folk-RNN) \cite{sturm16}. They can also receive conditional inputs like the encoding of the previous lines \cite{zhang14}. Richer architectures that combine models focusing on different properties can be used to generate more complex text, e.g., poetry based on pentameter and rhymes \cite{lau18}. Finally, sequence modeling can also be combined with reinforcement learning. For example, the authors of \cite{jaques16} use a Note-RNN model (based on single notes) trained using Deep Q-Network \cite{mnih15}; as rewards, they consider both the classic loss of sequence prediction models and a reward based on rules of music theory. In this way, the model learns a set of composition rules, while still maintaining information about the transition probabilities of the training data. The advantages of adopting an RL-based approach are described in Section \ref{rlbased}.

Due to the difficulties in working with long sequences, results in tasks like narrative generation are affected by a lack of coherence \cite{roemmele18}. Many approaches have been proposed to address this problem. For instance, stories can be generated in terms of events \cite{martin18} (i.e., tuples with subject, verb, object, and an additional \textit{wildcard}) by an encoder-decoder RNN (also known as Sequence-to-Sequence, see \cite{sutskever14}); events are modeled by another encoder-decoder RNN. Instead of events, it is also possible to focus on entities (i.e., vectors representing characters)~\cite{clark18}.

Sequence prediction models are also used for domains not commonly modeled as sequences, like images. Image modeling can be defined in a discrete way by means of a joint distribution of pixels: the model learns to predict the next pixel given all the previously generated ones. It starts at the top left pixel and then proceeds towards the bottom right. The two seminal architectures for sequence prediction of images are PixelRNN and PixelCNN \cite{oord16a}. The former is a two-dimensional RNN (based on rows or diagonals). The latter is a convolutional neural network (CNN) with an additional fixed dependency range (i.e., the convolution filters are masked in order to only use information about pixels above and to the left of the current one). To obtain better results, gated activation units can be used in place of rectified linear units between the masked convolutions; conditional inputs encoding high-level image descriptions can be used as well \cite{oord16b}. Notably, the Gated PixelCNN architecture can also be used for other types of data: WaveNet \cite{oord16c} implements it to generate audio based on the waveform, possibly guiding the generation with conditional inputs.

While intuitive in terms of architecture, RNNs are limited by the vanishing gradient problem and non-parallelizability in the time dimension \cite{le16}. Very recent works explore solutions to tackle these issues by means of structured state spaces \cite{gu22} and a combination of RNNs and Transformers \cite{peng23} (see Section \ref{transformers}).

\subsubsection{Applications}

As discussed, sequence prediction models have been used to learn to write poems or stories (by predicting a character, syllable, or word after the other); to compose music (by predicting a note or a waveform after the other); to draw images (by predicting a pixel after the other). In general, they can be used for any kind of time series forecasting \cite{lim21}. They can also be used for co-creativity, as in Creative Help \cite{roemmele18}.
Despite their simplicity, sequence prediction models are one of the most successful generative techniques. An interesting example is \textit{Sunspring}. It might be considered as the first AI-scripted movie: it was generated by a Char-RNN trained on thousands of sci-fi scripts \cite{miller19}. The quality of the result is demonstrated by the fact that it was able to reach the top ten at the annual Sci-Fi London Film Festival in its 48-Hour Film Challenge\footnote{Quite interestingly, the AI that wrote \textit{Sunspring} declared that its name was Benjamin, probably in honor of Walter Benjamin, the German philosopher who, already in 1935 \cite{benjamin68}, understood that new mechanical techniques related to art can radically change the public attitude to art and artists.}.

\subsubsection{Critical Discussion}

Sequence prediction models generate outputs that have characteristics of both exploratory and combinatorial creativity. They are based on probabilistic predictions and they are able to generate new outputs in the induced space, but they can also reuse sequences of tokens from different works, combining them together. There is no guarantee that the results will be valuable or novel, and classic methods such as RNNs lack surprise \cite{bunescu19}. It is worth noting that the possibility of using conditional inputs and being able to work at different levels of abstraction might indirectly lead to creative outputs, but creativity should then be attributed to the higher-level component (or human if the input is provided by the user) that is guiding the generation for specific elements and characteristics of the result.

\subsection{Transformer-Based Models} \label{transformers}

\subsubsection{Core Concepts}

Transformer-based models are neural networks based on the Transformer architecture \cite{vaswani17}. They represent the main example of foundation models \cite{bommasani21}, because of the leading role they have been assuming in language, vision, and robotics. A Transformer is an architecture for sequential modeling that does not require recurrent or convolutional layers. Instead, it only relies on a self-attention mechanism \cite{bahdanau15} that models long-distance context without a sequential dependency. Each layer consists of multi-head attention (i.e., several self-attention mechanisms running in parallel), a feed-forward network, and residual connections. Since self-attention is agnostic to token order, a technique called positional embedding is used to capture the ordering \cite{vaswani17}.

In principle, a Transformer is nothing more than an autoregressive model: it works by predicting the current token given the previous ones (see Section \ref{sequenceprediction}). However, few fundamental differences exist. A Transformer can also be trained by means of masked modeling: some of the input tokens are randomly masked, and the model has to learn how to reconstruct them from the entire context, and not only from the previous portions \cite{devlin18}. The possibility of dealing with very long sequences allows for prompting. By providing a natural language prompt in input, the model can generate the desired output, e.g., the answer to a question, a class between a set of classes for a given text, or a poem in a particular style \cite{brown20}. This is done by simply passing the prompt in input as a text, and then leveraging the model to predict what comes next (e.g., the answer to a question).
These advantages, together with the very large amount of data available, the increasing computational power, and the parallelism induced by their architecture, have led Transformer-based models to become the state of the art for several tasks. Nevertheless, the computational costs of the architecture from \cite{vaswani17} grow quadratically with the input size.

\subsubsection{Examples of Models}

Several Transformer-based approaches have been proposed in recent years. The design of specific Transformers for a variety of applications is presented in several surveys (e.g., \cite{bommasani21, khan22}) and books (e.g., \cite{tunstall22}).

The domain mostly influenced by Transformers is natural language processing (NLP). Bidirectional Encoder Representations from Transformers (BERT) \cite{devlin18} is a Transformer-based encoder trained for both predicting the next sentence (in an autoregressive fashion) and reconstructing masked tokens from the context. Several enhanced variations of the original model have been proposed, such as, for instance, solutions that remove the next-sentence pre-training objective \cite{liu19}, use inter-sentence coherence as an additional loss \cite{lan20}, or employ distillation \cite{hinton15} to train a smaller model \cite{sanh19}.
The other main approach is that used by the Generative Pre-trained Transformer (GPT) family \cite{radford18,radford19,brown20}. Here, a Transformer-based decoder is trained in an autoregressive way by additionally conditioning on the task of interest. After training, it can be used to perform a wide range of tasks by providing a description or a few demonstrations of the task. The effectiveness of this text-to-text generative approach has then been explored by T5 \cite{raffel20}. Many other large language models \cite{shoeybi19, touvron23, zhang22} have been proposed to achieve better results by means of more parameters and computation \cite{smith22}, or more qualitative data \cite{gunasekar23}. 
Mixture of Experts \cite{shazeer17} can be used as well in place of the feed-forward network to train a larger but lighter model (since only portions of it are used per task), as done by Generalist Language Model (GLaM) \cite{du22}. Finally, Bidirectional and Auto-Regressive Transformer (BART) \cite{lewis20} ideally merges together a BERT-encoder (trained by corrupting text with an arbitrary noising function) and a GPT-decoder (trained to reconstruct the original text autoregressively). Such an encoder-decoder architecture is able to achieve state-of-the-art results in machine translations, as well as other text-to-text tasks.

Transformer-based models have been used in domains different from language modeling (LM). Few have been proposed for music generation. One of the first examples was Music Transformer \cite{huang18}, which can generate one-minute music in Bach's style with internal consistency; another remarkable one is Musenet \cite{payne19}, which is able to produce 4-minute musical composition with a GPT-2 architecture; and, finally, it is worth mentioning Jukebox \cite{dhariwal20}, which can generate multiple minutes of music from raw audio by training a Sparse Transformer \cite{child19} (i.e., a Transformer with sparse factorization of the attention matrix in order to reduce from quadratic to linear scaling) over the low-dimensional discrete space induced by a VQ-VAE. Conditioning is always considered by means of genre, author, or instruments. MusicLM \cite{agostinelli23} additionally allows to generate music from text descriptions by aligning text and audio representation from different state-of-the-art models \cite{borsos22,huang22}.
Another important application domain is video-making. Video Vision Transformer (ViViT) \cite{arnab21} generates videos using classic Transformer architectures; Video Transformer (VidTr) \cite{zhang21} achieves state-of-the-art performance thanks to the standard deviation-based pooling method; and VideoGPT \cite{yan21} does so by learning discrete latent representations of raw video with VQ-VAE, and then training a GPT autoregressively. 

Transformers have been highly influential in computer vision too. The first model was Image Transformer \cite{parmar18}. It restricts the self-attention mechanism to attend to local neighborhoods, so larger images can be processed. Class-conditioned generation is also supported, by passing the embedding of the relative class in input. To avoid restricting self-attention to local neighborhoods, Vision Transformer~\cite{dosovitskiy21} divides an image into fixed-size patches, linearly embeds each of them, adds position embeddings, and then feeds the resulting sequence of vectors to a standard Transformer encoder. Masked Autoencoders (MAE) \cite{he22} instead uses an encoder-decoder architecture based on Transformers trained with masked image modeling (i.e., to reconstruct randomly masked pixels). A BERT adaptation to images called Bidirectional Encoder representation from Image Transformers (BEiT) \cite{bao22} has also been proposed. Masked image modeling has also been used together with classic autoregressive loss \cite{chen20}. Conversely, Vector Quantised-GAN (VQ-GAN) \cite{esser21} allows a Transformer to be based on vector quantization. A GAN learns an effective codebook of image constituents. To do so, the generator is implemented as an auto-encoder; vector quantization is applied over the latent representation returned by the encoder. It is then possible to efficiently encode an image in a sequence corresponding to the codebook indices of their embeddings. The Transformer is finally trained on that sequence to learn long-range interactions. These changes also allow us to avoid quadratic scaling, which is intractable for high-resolution images. Finally, DALL-E \cite{radford21} takes advantage of a discrete VAE. To generate images based on an input text, it learns a discrete image encoding; it concatenates the input text embedding with the image encoding; it learns autoregressively on them. CogView implements a similar architecture~\cite{ding21}.

Finally, Transformer-based models have also been used in multimodal settings, in which data sources are of different types. A survey can be found in \cite{suzuki22}. The first examples of these systems consider text and images together as the output of the Transformer architecture. By aligning their latent representations, images and texts can be generated by Transformer-based decoders given a multimodal representation. For instance,  Contrastive Language-Image Pre-training (CLIP) \cite{radford21} has an image encoder pre-trained together with a text encoder to generate a caption for an image. A Large-scale ImaGe and Noisy-text embedding (ALIGN) \cite{jia21}, based on similar mechanisms, can achieve remarkable performance through training based on a noisier dataset. In~\cite{tsimpoukelli21} the authors propose a frozen language model for multimodal few-shot learning: a vision encoder is trained to represent each image as a sequence of continuous embeddings, so that the frozen language model prompted with this embedding can generate the appropriate caption. In~\cite{fei22} the authors present Bridging-Vision-and-Language (BriVL), which performs multimodal tasks by learning from weak semantic correlation data. Finally, there is a trend toward even more complex multimodal models. For example, Video-Audio-Text Transformer (VATT) \cite{akbari21} learns to extract multimodal representations from video, audio, and text; instead, Gato \cite{reed22} serializes all data (e.g., text, images, games, other RL-related tasks) into a flat sequence of tokens that is then embedded and passed to a standard large-scale language model. Similarly, Gemini \cite{gemini23} achieves state-of-the-art performance in multimodal tasks by working on interleaved sequences of text, image, audio, and video as inputs; \cite{gemini24} extends it to Mixture-of-Experts setting. Finally, NExT-GPT \cite{wu23} handles any combination of four modalities (text, audio, image, and video) by connecting a language model with multimodal adaptors and diffusion decoders (see Section \ref{diffusion}).

\subsubsection{Applications}

Transformer-based large language models can be used for almost any NLP task, including text summarization, generation, and interaction. In order to do so, the model can be used as frozen (i.e., to provide latent representations in input to other models); can be fine-tuned for the specific objective; can be exploited with zero-shot, one-shot or few-shot setting by prompting the task or few demonstrations in input. Transfer learning can instead be used to perform image classification by means of Transformer-based models trained on images. Other domain-specific techniques can be used as well: for instance, PlotMachines \cite{rashkin20} learns to write narrative paragraphs not by receiving prompts, but by receiving plot outlines and representations of previous paragraphs.
From a generative learning perspective, Transformers have shown impressive performance in producing long sequences of texts and music or speech \cite{wang23}, as well as in generating images based on input text. 
Their application has not been limited to these data sources. For instance, AlphaFold uses a Transformer architecture to predict protein structure \cite{jumper21}; RecipeGPT employs it to generate recipes \cite{lee20}; and GitHub Copilot relies on it to support code development \cite{chen21}.

\subsubsection{Critical Discussion}

Given the fact that the Transformers can be considered as an evolution of sequence prediction models, the observations made for that class of models (see Section \ref{sequenceprediction}) apply also to them. However, the inherent characteristics of their architecture allow for larger models and higher-quality outputs, which are also able to capture a variety of dependencies of text across data sources. More in general, a broader conceptual space is induced. This means that domain-specific tasks might be addressed by means of solutions outside or at the boundary of the sub-space linked with that domain.
Moreover, possibly also through careful use of inputs (see Section \ref{inputbased}), their adoption might lead to transformational creativity. As far as Boden's criteria are concerned, there is no guarantee that the output of the Transformer architecture would be valuable, novel, or surprising, even though current state-of-the-art large language models (LLMs) achieve almost human-like performance in creative tests \cite{stevenson22,zhao24}.

\subsection{Diffusion Models} \label{diffusion}

\subsubsection{Core Concepts}

Diffusion models are a family of methods able to generate samples by gradually removing noise from a signal \cite{sohldickstein15}. The most representative approach is the Denoising Diffusion Probabilistic Model (DDPM) \cite{ho20}. An input $\mathbf{x_0}$ is corrupted by gradually adding noise until obtaining an $\mathbf{x_T}$ from a pre-defined distribution; the model then has to reverse the process. Each timestep $t$ corresponds to a certain noise level; $\mathbf{x_t}$ can be seen as a mixture of $\mathbf{x_0}$ with some noise $\mathbf{\epsilon}$ whose ratio is determined by $t$. The model learns a function $\epsilon_{\theta}$ to predict the noise component of $\mathbf{x_t}$ by minimizing the mean-squared error. $\mathbf{x_{t-1}}$ is then obtained from a diagonal Gaussian with mean as a function of $\epsilon_{\theta}\!\left(\mathbf{x_t},t\right)$, and with a fixed \cite{ho20} or learned \cite{nichol21a} variance. In other words, it learns to associate points from a predefined random distribution with real data through iterative denoising. Because of this, at inference time, a diffusion model can generate a new sample by starting from pure random noise. The generation can also be conditioned by simply modifying the noise perturbation so that it depends on the conditional information. However, this iterative sampling process might potentially lead to slow generation; a proposed solution is to induce self-consistency, i.e., ensuring that points on the same trajectory map to the same initial ones \cite{song23}. In this way, the output can be obtained in a single step.

The aforementioned diffusion process is similar to the one followed by score-based generative models \cite{song19,song20}. Instead of noise, here a model is trained to learn the score, i.e., the gradient of the log probability density with respect to real data. The samples are then obtained using Langevin dynamics \cite{welling11}. Despite the differences, both of them can be seen as specific, discrete cases of Stochastic Differential Equations \cite{song21}.

\subsubsection{Examples of Models}

Diffusion models have been primarily used for image generation. In order to generate higher-quality images and to allow text-to-image generation, a variety of effective methods for conditioning have been proposed. 
A possibility is to use classifier guidance \cite{dhariwal21}: the diffusion score (i.e., the added noise) includes the gradient of the log-likelihood of an auxiliary classifier model. An alternative is classifier-free guidance \cite{ho21a}: to avoid learning an additional model, a single neural network is used to parameterize two diffusion models, one conditional and one unconditional; the two models are then jointly trained by randomly setting the class for the unconditional model. Finally, the sampling is performed using a linear combination of conditional and unconditional score estimates. Guided Language to Image Diffusion for Generation and Editing (GLIDE) \cite{nichol22} demonstrates how classifier-free guidance can be effectively used to generate text-conditional images. In addition, it shows how diffusion models can be used for image editing by fine-tuning in order to reconstruct masked regions. Performance improvement can be obtained by means of a cascade of multiple diffusion models performing conditioning augmentation \cite{ho22b}. Notably, the diffusion model can operate on latent vectors instead of real images. Stable Diffusion \cite{rombach22} employs a diffusion model in the latent space of a pre-trained autoencoder. Similarly, DALL-E 2 \cite{ramesh22} generates images by conditioning with image representations. At first, it learns a prior diffusion model to generate possible CLIP image embeddings from a given text caption, i.e., conditioned by its CLIP text embedding. Then, a diffusion decoder produces images conditioned by the image embedding. The generation quality can be further improved by means of generated captions for the images in the training set \cite{betker23}. Imagen \cite{saharia22} uses instead a cascaded diffusion decoder, together with a frozen language model as a text encoder to increase the quality of output.

Although the approach is particularly suitable for images, applications to other data sources have been developed as well. DiffWave \cite{kong21} and WaveGrad \cite{chen21} use diffusion models to generate audio. They overcome the continuous-discrete dichotomy by working on waveform. Another possibility is to use an auto-encoder like MusicVAE \cite{roberts18} to transform the sequence into a set of continuous latent vectors, on which training a diffusion model \cite{mittal21}. Resembling image generators, Contrastive Language-Audio Pretraining (CLAP) embeddings \cite{elizalde22} can be used to generate audio by conditioning on text descriptions \cite{liu23}. Diffusion-LM \cite{li22} employs diffusion models to write text by denoising a sequence of Gaussian vectors into continuous word vectors (then converted into discrete words by a rounding step); DiffuSeq \cite{gong23} performs sequence-to-sequence generation tasks by embedding source and target sequences in the same embedding space through a Transformer architecture. Diffusion models have been used for 3D generation as well \cite{nichol22b}. Finally, diffusion models for video have also been proposed, based on gradient-based conditioning \cite{ho22}, and on processing latent spacetime patches. In particular, with respect to the latter, Sora \cite{brooks24} first turns videos into sequences of patches and then uses a diffusion Transformer to predict the original patches from random noise (and conditioning inputs like text prompts), improving sample quality and flexibility.

\subsubsection{Applications}

Despite their recent introduction, diffusion models have been used to generate audio, music, and video, as well as to generate and edit images conditioned on input text, e.g., with in-painting \cite{lugmayr22} or subject-driven generation \cite{ruiz23}; we refer to \cite{yang23} for a comprehensive survey of this area.
Indeed, they lead to higher-quality outputs than the previous state-of-the-art models. In particular, DALL-E 2 and Stable Diffusion have been able to produce images from textual instructions with superior fidelity and variety.

\subsubsection{Critical Discussion}

Diffusion models learn a mapping between real images and a Gaussian latent space. Because of this, they are an example of exploratory creativity: they randomly sample from that space, and then they possibly navigate it in the direction imposed by conditional inputs. There is no guarantee that the results will be valuable, novel, or surprising, even though these approaches are able to generate outputs characterized by a high variety. As already argued, novelty and surprise may only arise due to the conditioning input (for example, a human describing a novel combination of elements), i.e., the model is not imaginative on its own.

\subsection{Reinforcement Learning-Based Methods} \label{rlbased}

\subsubsection{Core Concepts}

With RL-based methods, we aim to indicate all the generative models whose training relies on maximizing a reward. These models are based on the architectures introduced so far, e.g., they can be GANs or autoregressive models. The difference is that they are not (only) trained to fool the discriminative part or to reduce prediction error. The typical framework considers the generative model as the agent; each action causes a modification to the current product, i.e., the state; and the agent learns a policy that maximizes the cumulative reward. Therefore, the reward can be used to impose desired behavior on the generative model. The RL-based approach can be implemented for the entire training or for fine-tuning a pre-trained model. The sampling scheme remains the same and depends on the chosen generative model.

\subsubsection{Examples of Models}

A first example is Objective-Reinforced GAN (ORGAN) \cite{guimaraes17}.
Here, RL is used not only to adapt GANs to sequential tasks but also to provide additional learning signals, such as rewards from specific-domain objectives (e.g., tonality and ratio of steps for music generation). Also \cite{yi18} follows this path by using rewards like fluency, coherence, meaningfulness, and overall quality to generate poems. Another possibility is to use the metrics used at test time (e.g., BLEU or ROUGE) \cite{bahdanau17,ranzato16}. Instead, RL-DUET \cite{jiang20} casts online music accompaniment generation as an RL problem with an ensemble of reward models, i.e., autoregressive models trained with or without the whole context, and with or without the human-produced counterpart. In this way, inter-coherence (between humans and machines) and intra-coherence can be obtained. Finally, Intelli-Paint \cite{singh22} can paint in human style by using a sequential planner that learns a painting policy to predict vectorized brushstroke parameters from the current state.

RL can also be used to fine-tune a pre-trained generative model. Doodle-SDQ \cite{zhou18} first learns to draw simple strokes using supervised learning; then, it improves its generation by means of rewards about similarity, color, and line movement. Conversely, the authors of \cite{tambwekar19} suggest to consider a pre-trained LSTM language model as a policy model. Fine-tuning then aims at maximizing the probability that a given event occurs at the end of the narrative. RL Tuner \cite{jaques17a} uses RL to fine-tune a Note-RNN \cite{eck02} to produce music that follows a set of music theory rules. To avoid forgetting note probabilities learned from the data, the probability value returned by a copy of the pre-trained Note-RNN can be used as an additional reward. Sequence Tutor \cite{jaques17b} generalizes this idea of learning a policy that trades off staying close to the data distribution while improving performance on specific metrics. A comprehensive critical discussion of the rewards for RL-based generative models can be found in \cite{franceschelli24}.
Finally, RL can be used to help models follow human preferences \cite{christiano17} or feedback. The latter technique is referred to as Reinforcement Learning from Human Feedback (RLHF) \cite{stiennon20}. For example, ChatGPT \cite{openai22} is an interactive version of GPT-3 \citep{brown20} (initially) and GPT-4 \citep{openai23} (at the time of writing), fine-tuned to maximize a learned reward of human values. RLHF improves its conversational skills while mitigating mistakes and biases; because of this, it has become a standard de facto for fine-tuning large language models, e.g., in \cite{touvron23b}. It is also possible to use AI and not human feedback \cite{bai22b,lee23b}. However, RLHF has some limitations \cite{casper23}, and alternative RL-free strategies are increasingly popular (e.g., \cite{rafailov23}).

\subsubsection{Applications}

As seen, RL-based models can be used to fully train or fine-tune generative models for different tasks; ideally, for any task that could benefit from domain-specific objectives. This is the case of music and molecule generation \cite{guimaraes17,jaques17b}, but also of dialogue generation \cite{li16}, image generation \cite{black23} and painting \cite{huang19}. In addition, the sequential nature of RL can help as well in all the tasks requiring to deal with new directives during generation (e.g., music interaction). Finally, RLHF can be used to directly optimize models for creative tasks, e.g., poetry \cite{pardinas23}.

\subsubsection{Critical Discussion}

The evaluation of the creativity of RL-based models depends on how the agent is implemented and which rewards are considered. The learned space of solutions depends on the used rewards (and on the pre-training technique in case of fine-tuning). They typically contain an adversarial signal or a likelihood with respect to the training data; thus, combinatorial or exploratory creativity is obtained. However, additional rewards can have the effect of transforming that space (see Section \ref{creativityoriented}). As far as Boden's criteria are concerned, value is typically ensured by some qualitative domain-specific reward or by approaching human preferences. This also allows us to consider them as appreciative, as discussed in Section \ref{gandiscussion}. Novelty and surprise might be achieved as well by means of specific rewards; however, this is not the case for current models.

\subsection{Input-Based Methods} \label{inputbased}

\subsubsection{Core Concepts}

The last class of methods we consider in our analysis is not about a different generative model. On the contrary, it is about a different way to sample results from (pre-trained) generative models, namely by means of its inputs. Two different approaches can be used. The first is about carefully selecting or optimizing the input to a generative model (e.g., the latent vector or the text prompt) so that to obtain the desired output. The second approach is about optimizing the input so that it directly becomes the desired output. They rely on losses that are usually based on features learned by neural networks. While the two approaches are technically different, both of them aim at obtaining better outputs by exploiting the knowledge of the pre-trained model through the optimization of the inputs.

\subsubsection{Examples of Models}

The first approach consists of carefully modifying the input of a generative model until the output matches the desired properties. The main example is VQGAN-CLIP \cite{crowson22}. Given a text description, VQGAN produces a candidate image from a random latent vector; the vector is then optimized by minimizing the distance between the embeddings of the description and the candidate image. Both embeddings are computed using CLIP \cite{radford21}. Variants can be implemented as in Wav2CLIP \cite{wu22}, where an audio encoder is learned to match the CLIP encoders so that VQGAN-CLIP can be used from raw audio; or as in music2video \cite{jang22}, where videos are generated from audio a frame after the other by both minimizing the distance between subsequent frames, and the distance between image and music segment embedded by Wav2CLIP. In addition to the random latent vector, the text or audio description can be optimized as well. This can be performed by the users through many iterations of careful adjustments, or by means of an automated procedure. The latter is commonly known as prompt tuning. Prompt tuning is about producing prompts via backpropagation; the optimized prompts can then condition frozen language models in order to perform specific tasks without having to fine-tune them \cite{lester21}. An additional model can also be trained to output the desired prompt \cite{levine22}. Finally, image generators such as VQGAN can also be exploited in other ways, i.e., with binary-tournament genetic algorithm \cite{fernando21} or more complex evolution strategies \cite{tian22}. Another possibility is to optimize the input so that the generated output maximizes a target neuron of an image classifier \cite{nguyen16}. This helps generate what that neuron has learned. The desired latent vector can also be produced by an additional model \cite{nguyen17}.

The second approach consists of optimizing the inputs to transform them into the desired outputs. DeepDream \cite{mordvintsev15} generates ``hallucinated'' images by modifying the input to maximize the activation of a certain layer from a pre-trained classifier. Artistic style transfer is based on the same idea. Given an input image and a target image, the former is modified by means of both style and content losses thanks to a pre-trained classifier. The content loss is minimized if the current and the original input images generate the same outputs from the hidden layers. The style loss is minimized if the current and target images have the same pattern of correlation between feature maps in the hidden layers \cite{gatys16}. Control over results can be improved by considering additional losses about color, space, and scale \cite{gatys17}.

\subsubsection{Applications}

Input-based methods can be used with any generative model to produce the desired output. With language models, they can exploit their generality in several specific tasks without fine-tuning them. For instance, prompt tuning can be used by writers for co-creation \cite{chakrabarty23} or to force LLMs to \textit{brainstorm} \cite{summers23}. With image generators, they can obtain drawings adherent to given descriptions, or high-quality but yet peculiar paintings like colorist \cite{fernando21}, abstract \cite{tian22} or alien \cite{snell21} artworks. We believe applications to other domains are yet to come. Both types of input-based methods can be used not only to produce desired outputs or to transfer styles; they can also be used to better analyze what is inside the network \cite{nguyen16,olah17}.

\subsubsection{Critical Discussion}

Since input-based methods are applied to pre-trained generative models, the space of solutions in which they work is the one induced by those models, i.e., the common spaces we can derive from real data. Nonetheless, some techniques may be able to cause productions that are outside that space or at its boundaries, i.e., to cause transformational creativity. This might happen if the model is general, and the output for a specific task is not only sampled from the sub-space of solutions for that task (e.g., with prompt tuning over a language model).
Input-based methods are also valuable: the input optimization itself is typically guided by some sort of qualitative loss. On the other hand, they are not explicitly novel or surprising (although the results might seem so). However, nothing prevents optimizing the loss in such directions (see Section \ref{creativityoriented}).

\subsection{Practical Assessment of Creativity-Oriented Methods} \label{creativityoriented}

We conclude this analysis of generative models with a discussion of how they might increase their \textit{creativity} according to Boden's definition. We have discussed how the presence of a recognition model (e.g., a discriminative model or a reward model) helps ensure the value of the products. In the same way, novelty and surprise can be fostered by the integration of other components.
A straightforward way to obtain novel and surprising outputs is to train a generative model by means of novelty and surprise objectives. This is the core idea behind Creative Adversarial Network (CAN) \cite{elgammal17,sbai19}. In addition to the classic discriminative signal, i.e., a value loss, the generator is also trained to optimize a novelty loss. This is defined as the deviation from style norms, i.e., the error related to the prediction of the style of the generated image. The sum of the two training signals helps the model learn to produce artworks that are different (in style) from the training data. The same approach has been used to develop a creative StyleGAN, i.e., StyleCAN \cite{jha21}. 
Another, very simple way to augment the training signal of a generative model with \textit{creativity-oriented} objectives is by means of RL-based methods.
The choice of the reward structure is the fundamental element in the design of effective generative reinforcement learning systems. Rewards should teach the model to generate an output with a high level of novelty and surprise. An example is ORGAN \cite{guimaraes17}, where appropriate reward functions can be used. For instance, statistical measures (e.g., Chi-squared) or metrics of distance between distributions (e.g., KL divergence) might be used to ground ideas of novelty and surprise.

Another possibility is the development of an input-based method where the input is optimized to obtain a product that is valuable, novel, and surprising. This may be achieved either by forcing a further exploration of the latent space (e.g., by means of evolutionary search \cite{fernandes20}), or by defining appropriate loss functions to perform gradient descent over the input. All these methodologies are also called \textit{active divergence} \cite{berns20} since they aim to generate in ways that do not simply reproduce training data. A survey on active divergence can be found in \cite{broad21}. A different output can also be obtained by carefully altering the probability distribution of the model, e.g., by scaling its probabilities with learned functions to maximize target properties \cite{dathathri20,snell23,yang21}.

A different approach is followed by the Composer-Audience architecture \cite{bunescu19}. Two models are considered: the Audience, a simple sequence prediction model trained on a given dataset; the Composer, another sequence prediction model trained on a different dataset. In addition, the Composer also receives the next-token expectations from the Audience, and it learns when to follow its guidance and when to diverge from expectations, i.e., when to be surprising. For instance, it can learn to produce jokes by considering non-humorous texts to train the Audience, and humorous texts to train the Composer. Even though this approach is useful for learning how to generate valuable and surprising output, it is only applicable when paired datasets are available.

\begin{table}[t]
    \centering
        
    \begin{tabular}{cccc}
        \toprule
        Generative family & Type of creativity & Boden's criteria & Creative suggestions \\
        \midrule
        VAE & Exploratory & \specialcell[l]{$\pmb{\sim}$ Value \\ $\pmb{\sim}$ Novelty \\ $\pmb{\sim}$ Surprise} & \specialcell{Creativity-oriented \\ input-based methods} \\ 
        \cmidrule(r){1-4}
        GAN & Exploratory & \specialcell[l]{$\pmb{\checkmark}$ Value \\ $\pmb{\sim}$ Novelty \\ $\pmb{\sim}$ Surprise} & \specialcell{CAN; \\ Creativity-oriented \\ input-based methods} \\
        \cmidrule(r){1-4}
        \specialcell{Sequence \\ prediction \\ model} & \specialcell{Combinatorial, \\ Exploratory} & \specialcell[l]{$\pmb{\sim}$ Value \\ $\pmb{\sim}$ Novelty \\ $\pmb{\times}$ Surprise} & \specialcell{Composer-Audience; \\ Creativity-oriented \\ RL-based methods} \\ 
        \cmidrule(r){1-4}
        \specialcell{Transformer-\\ based \\ models} & \specialcell{Combinatorial, \\ Exploratory, \\ Transformational} & \specialcell[l]{$\pmb{\sim}$ Value \\ $\pmb{\sim}$ Novelty \\ $\pmb{\times}$ Surprise} & \specialcell{Creativity-oriented \\ prompt tuning or \\ RL-based methods} \\
        \cmidrule(r){1-4}
        \specialcell{Diffusion \\ models} & Exploratory & \specialcell[l]{$\pmb{\sim}$ Value \\ $\pmb{\sim}$ Novelty \\ $\pmb{\sim}$ Surprise} & \specialcell{Creativity-oriented \\ input-based methods} \\
        \cmidrule(r){1-4}
        \specialcell{RL-based \\ methods} & \specialcell{Combinatorial, \\ Exploratory, \\ Transformational} & \specialcell[l]{$\pmb{\checkmark}$ Value \\ $\pmb{\sim}$ Novelty \\ $\pmb{\sim}$ Surprise} & \specialcell{Intrinsic rewards; \\ Novelty-based rewards} \\
        \cmidrule(r){1-4}
        \specialcell{Input-based \\ methods} & \specialcell{Exploratory, \\ Transformational} & \specialcell[l]{$\pmb{\checkmark}$ Value \\ $\pmb{\sim}$ Novelty \\ $\pmb{\sim}$ Surprise} & \specialcell{Evolutionary search; \\ Novelty-based optimization} \\
        \bottomrule
    \end{tabular}
    \caption{Summary of all the methods explained so far, considering their type of creativity as discussed in the corresponding subsections; the possible presence of Boden's criteria ($\pmb{\checkmark}$ if induced by the training process; $\pmb{\sim}$ if not considered; $\pmb{\times}$ if excluded); and some practical suggestions to achieve a higher degree of creativity.}
    \label{tab:table_generative_models}
\end{table}

As far as the type of creativity is concerned, there can be ways to achieve a better exploration or even transformation of the space of solutions. For example, since CAN novelty loss is used during training, it learns to diverge from the distribution of real data. The same is true for RL-based methods with novelty and surprise rewards (especially if the training happens from scratch). Finally, a more explored or transformed space may be reached using RL-based methods driven by curiosity \cite{burda19}: an agent can learn to be creative and discover new patterns thanks to intrinsic rewards to measure novelty, interestingness, and surprise. This can be done by training a predictive model of the growing data history and by using its learning progress as the reward. In this way, the agent is motivated to make things the predictor does not yet know. If an external qualitative reward is considered as well, the agent should in theory learn to do things that are new, but still valuable \cite{schmidhuber10a}. The same idea can also be applied to different techniques like evolutionary strategies \cite{machado07}. Deep Learning Novelty Explorer (DeLeNoX) \cite{liapis13a} uses a denoising autoencoder to learn low-dimensional representations of the last generated artifacts. Then, a population of candidate artifacts (in terms of their representation) is evolved through a feasible-infeasible novelty search \cite{liapis13b} in order to maximize the distances between them, i.e., to increase their novelty, while still considering qualitative constraints. Other evolutionary strategies might be considered as well to search the space of artifacts for novel \cite{lehman11} and surprising \cite{gravina16} results. Instead of relying on manually crafted metrics, Quality Diversity through Human Feedback (QDHF) \cite{li23} uses human feedback for computing quality and distance in learned latent projection for computing diversity. Quality-Diversity through AI Feedback (QDAIF) \cite{bradley23} makes the model more independent in searching and innovating by totally relying on its own feedback for both quality and diversity.

Table \ref{tab:table_generative_models} summarizes all the generative approaches discussed in this section, highlighting their characteristics from a machine creativity perspective.

\section{Creativity Measures} \label{creativitymeasures}

In this section, we present different methodologies to evaluate the creativity of artifacts generated by artificial agents. These can typically be extended to human-generated artifacts. For each of them, we explore the core concepts, the dimensions of creativity that are considered, the evaluation protocol, and, finally, we critically assess them.
The presence of several different proposals can be associated with the fact that it is not always straightforward to determine the ‘‘right’’ question to ask in an evaluation of a creative artifact \cite{gatt17}. For instance, the Generation, Evaluation, and Metrics (GEM) Benchmark for natural language generation \cite{gehrmann21} does not contain any creativity evaluation measure. This is due to the inadequacy of the creative metrics proposed to date to correctly capture and measure all the necessary dimensions of creativity. The focus of our overview is on measures that are based on (or associated with) machine learning techniques. It is worth noting that some of them might be calculated without using machine learning, but we will refer to an implementation based on the latter. For an in-depth overview of creativity measures not strictly related to machines, we refer to \cite{said17}.
Table \ref{tab:table_creativity_measures} reports all the evaluation methods considered in this section, highlighting the dimensions they try to capture, their applicability, and their limitations. We will discuss these aspects in the remainder of this section.

\begin{table}[htp]
    \centering
    
    \begin{tabular}{ccccl}
        \toprule
        Name & What evaluates & How evaluates & Applicability & \centercell{Limits}
        \\
        \midrule
        \specialcell{Lovelace 2.0 \\ Test} & \specialcell{Evaluators' \\creativity definition} & \specialcell{Mean number of \\challenges per evaluator} & General & \specialcell[l]{
            \tabitem Requires a substantial\\ human intervention} \\ \cmidrule(r){1-5}
        \specialcell{Ritchie's \\ criteria} & \specialcell{Quality, \\ novelty, \\typicality} & \specialcell{Human opinions \\ (then elaborated \\ through 18 criteria)} & General & \specialcell[l]{
            \tabitem Requires human evaluation \\
            \tabitem Requires to state thresholds \\
            \tabitem No innovation definition} \\ \cmidrule(r){1-5}
        FACE & \specialcell{Tuples of \\ generative \\ acts} & \specialcell{Volume of acts, number \\ of acts, quality (through\\ aesthetic measure)} & General & \specialcell[l]{
            \tabitem Abstract method \\ 
            \tabitem Definition of aesthetic \\ measure left to the user} \\ \cmidrule(r){1-5}
        SPECS & \specialcell{What we state \\ creativity is} & \specialcell{Identification and test \\ of standards for the \\ creativity components} & General & \specialcell[l]{
            \tabitem More a framework for eva-\\luation method definition \\ than a real method} \\ \cmidrule(r){1-5}
        \specialcell{Creativity \\ implication \\ network} & \specialcell{Value,\\ novelty} & \specialcell{Similarity between works \\ (considering subsequent \\ works for value and pre-\\vious works for novelty)} & General & \specialcell[l]{
            \tabitem Not possible to accurately\\ measure the creativity of\\
            the most recent works \\ 
            \tabitem Wrong creativity and \\ time-positioning correlation} \\ \cmidrule(r){1-5}
        \specialcell{Chef Watson \\ (assessment part)} & \specialcell{Novelty,\\ quality} & \specialcell{Bayesian surprise, smell \\ pleasantness regression} & \specialcell{Specific \\ (recipes)} & \specialcell[l]{
            \tabitem Requires human ratings \\ of pleasantness} \\ \cmidrule(r){1-5}
        DARCI & \specialcell{Art \\ appreciation} & \specialcell{Neural network to \\ associate image featu-\\ res and description} & \specialcell{Specific \\ (visual art)} & \specialcell[l]{
            \tabitem Not based on product \\
            \tabitem Considers just one of \\ the creative tripod} \\ \cmidrule(r){1-5}
        \specialcell{PIERRE - \\ Evaluation part} & \specialcell{Novelty,\\ quality} & \specialcell{Count of new combi-\\nations; user ratings} & \specialcell{Specific \\ (recipes)} & \specialcell[l]{
            \tabitem Requires user ratings \\ over ingredients} \\ \cmidrule(r){1-5}
        \specialcell{EVE'} & \specialcell{Feelings,\\ meanings} & \specialcell{Negative log of prediction \\ and posterior probability} & \specialcell{General} & \specialcell[l]{
            \tabitem Requires a way to explain \\
            \tabitem Value only through meaning} \\ \cmidrule(r){1-5}
        \specialcell{Common model \\ of creativity \\ for design} & \specialcell{Novelty, \\ value, \\ surprise} & \specialcell{K-Means on a description \\ space and a performance \\ space; degree of violation \\ of anticipated patterns} & \specialcell{Specific \\(design)} & \specialcell[l]{
            \tabitem Requires to define \\ attribute-value pairs \\
            \tabitem Requires to define \\ clustering parameters} \\ \cmidrule(r){1-5}
        Unexpectedness & \specialcell{Novelty, sur-\\prise, trans-\\formativity} & \specialcell{Possibility to update,\\ and degree of violation \\ of expectations} & General & \specialcell[l]{
            \tabitem Does not take care of value} \\ \cmidrule(r){1-5}
        \specialcell{Essential criteria \\ of creativity} & \specialcell{Value, \\ novelty, \\ surprise} & \specialcell{Sum of performance va-\\ riables, distance between\\ artifacts and between re-\\ al and expected artifact} & General & \specialcell[l]{
            \tabitem Requires to define \\ performance variables \\
            \tabitem Requires to define \\ clustering parameters} \\ \cmidrule(r){1-5}
        \specialcell{Computational \\ metrics for \\ storytelling} & \specialcell{Novelty, rarity, \\ recreational \\ effort, surprise} & \specialcell{Distance between do-\\minant terms, consecutive \\ fragments/clusters of terms} & \specialcell{Specific \\ (storytelling)} & \specialcell[l]{
            \tabitem Requires to define domination \\
            \tabitem Requires to define \\ clustering parameters} \\
        \bottomrule
    \end{tabular}
    
    \caption{Summary of creativity evaluation methods and their characteristics.}
    \label{tab:table_creativity_measures}
    
\end{table}

\subsection{Lovelace 2.0 Test}

\subsubsection{Overview}

The Lovelace Test (LT) \cite{bringsjord01} was proposed in 2001 as a \textit{creativity-oriented} alternative to the world-famous Turing test \cite{turing50}. More formally, LT is defined as follows:
\begin{definition}
An artificial agent $A$, designed by $H$, passes LT if and only if:

1) $A$ outputs $o$;

2) $A$'s outputting $o$ is not the result of a fluke hardware error but of processes $A$ can repeat;

3) $H$ (or someone who knows what $H$ knows, and has $H$'s resources) cannot explain how $A$ produced $o$ by appealing to $A$'s architecture, knowledge-base, and core functions.
\end{definition}

LT provides several insights for understanding and quantifying machine creativity, but it is rather abstract. For these reasons, a 2.0 version has been proposed \cite{riedl14}. The so-called Lovelace 2.0 Test is defined as:
\begin{definition}
Artificial agent $A$ is challenged as follows:

1) $A$ must create an artifact $o$ of type $t$;

2) $o$ must conform to a set of constraints $C$ where $c_i \in C$ is any criterion expressible in natural language;

3) a human evaluator $h$, having chosen $t$ and $C$, is satisfied that $o$ is a valid instance of $t$ and meets $C$;

4) a human referee $r$ determines the combination of $t$ and $C$ to not be unrealistic for an average human.
\end{definition}

\subsubsection{Dimensions of Creativity Considered}

Since the evaluation depends on the tests performed by human evaluators, the dimensions of creativity considered by it might vary greatly. This allows for considering value, novelty, and surprise, as well as domain-specific dimensions.

\subsubsection{Protocol for Evaluation}

The Lovelace 2.0 Test can be used to quantify the creativity of an artificial agent - by means of its artificial productions - considering a set $H$ of human evaluators. With $n_i$ as the first test performed by evaluator $h_i \in H$ not passed by the agent, the creativity of the artificial agent can be expressed as the mean number of challenges-per-evaluator passed: $\sum_i \frac{\left(n_i\right)}{\left|H\right|}$.

\subsubsection{Critical Examination}

This methodology represents an effective way to measure creativity since it is ideally applicable to any field and it is quantitative. Although the latter is not based on machine learning, it may be used in principle for performing (some of) the tests. However, this methodology requires considerable human intervention in order to define all the tests.

\subsection{Ritchie's Criteria}

\subsubsection{Overview}

Ritchie's Criteria \cite{ritchie07} is a set of criteria for evaluating the extent to which a program has been creative (or not) in generating artifacts. These criteria are based on three main factors: novelty, quality, and typicality. \cite{ritchie07} contains a proposed series of criteria, but according to the authors they should only be intended as a ``repertoire''.

\subsubsection{Dimensions of Creativity Considered}

Ritchie's Criteria are based on three factors: quality, typicality, and novelty. Quality measures how much an item is a high-quality example of its genre. Typicality measures how much an item is an example of the artifact class in question. Novelty measures the dissimilarity of an item with respect to existing examples in its class. Quality and typicality are collected using human opinions about the produced artifacts or using ``measurable'' characteristics about, for instance, syntax and metric (for poetry generator). On the other hand, novelty is intended as the sum of ``untypicality'' (the opposite of typicality) and innovation.

\subsubsection{Protocol for Evaluation}

The computation of the criteria is based on the analysis of the result set of produced artifacts, along with the inspiring set (composed by artifacts of that field used during training and/or generation). It also requires the definition of quality and typicality indicators for the artifacts considered.
More specifically, the proposed criteria are: the average of typicality or quality over the result set; the proportion of items with good typicality score, which is also characterized by high quality; the proportion of the output that falls into the category of untypical but high-valued; the ratio between untypical high-valued items and typical high-valued items; the proportion of the inspiring set that has been reproduced in the result set.
The assessment of these criteria is performed on both the entire result set and on the subset that does not contain any item from the inspiring set.

\subsubsection{Critical Examination}

These measures represent a promising way to evaluate creativity, but their application is not straightforward. In fact, they do not clearly specify how to measure novelty in terms of innovation. Furthermore, all measures require a large number of thresholds to be set (and the results are very sensitive to such thresholds \cite{pereira05}). The criteria for the selection of these thresholds are not trivial per se. It is difficult to identify a general methodology for setting these values. Finally, the collection of correct human opinions (in terms of consistency of measurement methodology, audience selection, etc.) is not a trivial task either.

\subsection{FACE}

\subsubsection{Overview}

In \cite{colton11} the authors introduce FACE as a descriptive model of the creative act of artificial systems. A creative act is considered a non-empty tuple of generative acts. FACE is designed to provide assessors with both quantitative and qualitative evaluations.

\subsubsection{Dimensions of Creativity Considered}

While FACE can be used to evaluate a product as the consequence of a creative act, its focus is on the process. The qualitative evaluation is therefore left to the aesthetic function; the dimensions considered depend in turn on how it has been defined (or generated).

\subsubsection{Protocol for Evaluation}

More specifically, the FACE model considers a creative act as a non-empty tuple of generative acts of eight possible types: an expression of a concept, i.e., an instance of an \textit{(input, output)} pair produced by the system; a method for generating expressions of a concept; a concept; a method for generating concepts; an aesthetic measure; a method for generating aesthetic measures; an item of framing information, i.e., a piece of additional information or description regarding the generative act; a method for generating framing information. It is then possible to use it in a quantitative way (how many acts are produced); in a cumulative way (how many types of acts are considered); and in a qualitative way (by means of the aesthetic measure taken into consideration).

\subsubsection{Critical Examination}

The FACE model represents a very comprehensive set of concepts and methodologies for assessing machine creativity. However, one of the most challenging aspects of FACE is the definition of the aesthetic measure, which is not specified; potentially, it might be defined by the system itself, counting as a potential creative act. This may award systems performing self-evaluation, i.e., guiding their generation based on learned objectives. This in theory might mean the systems are incentivized to develop their own tastes, which is an important part of human creativity.

\subsection{SPECS}

\subsubsection{Overview}

The Standardized Procedure for Evaluating Creative Systems (SPECS) \cite{jordanous12} is a framework for the evaluation of creative systems, which can easily be adapted to many different potential domains. The framework is based on the definition of fourteen ``components'' used to evaluate machine creativity.  

\subsubsection{Dimensions of Creativity Considered}

The fourteen key components of SPECS are: active involvement and persistence; dealing with uncertainty; domain competence; general intellect; generation of results; independence and freedom; intention and emotional involvement; originality; progression and development; social interaction and communication; spontaneity/subconscious processing; thinking and evaluation; value; and variety, divergence, and experimentation. However, SPECS does not restrict researchers to use all of them; moreover, domain-specific components can be added as well.

\subsubsection{Protocol for Evaluation}

SPECS is composed of three steps. The first one requires providing a definition of creativity that the system should satisfy, using the suggested components, and potentially other domain-specific ones. The second requires to specify the standards for evaluating such components. The third requires to test the system against the standards and report the results.

\subsubsection{Critical Examination}

This is an effective framework for working with computational creativity, but it cannot be considered as a practical evaluation method. Its effectiveness is strongly dependent on which components are considered and how they are evaluated for each specific task. In \cite{jordanous14} the author discusses how SPECS satisfies Ritchie's criteria and it is more comprehensive and expressive than the FACE model and the creative tripod \cite{colton08} (according to which a creative system should exhibit skills, appreciation, and imagination) in a meta-evaluation test. They also use a human opinion survey based on five criteria: correctness, usefulness, faithfulness (as a model of creativity), usability (of the methodology), and generality. 
SPECS is evaluated considering music improvisation generators; it is also judged by the developers of the generative systems. In particular, the authors show that SPECS can help obtain additional insights into how a generative model works and how it can be improved.

\subsection{Creativity Implication Network}

\subsubsection{Overview}

A different method of quantifying creativity is based on building an art graph called Creativity Implication Network \cite{elgammal15}. Given a set of artworks, a directed graph can be defined by considering a vertex for each artwork. More specifically, an arc connecting artwork $p_i$ to $p_j$ is inserted if $p_i$ has been created before $p_j$. A positive weight $w_{ij}$ quantifying the similarity score between the two artworks under consideration is associated with each arc. The creativity of an artwork is then derived by means of computations on the resulting graph.

\subsubsection{Dimensions of Creativity Considered}

This method captures both value and novelty. The value is defined as the influence on future artworks. The novelty is defined as the dissimilarity between the artwork and the previous ones.

\subsubsection{Protocol for Evaluation}

The derivation of the Creativity Implication Network requires a similarity function to compute the similarity scores; its definition is left to the researchers, but it can be based on ML techniques (as in the original paper, where computer vision techniques were used).
Given the network, the creativity of artwork $p_i$ depends on the similarity with the previous artworks (higher the similarity, lower the creativity) and with the subsequent artworks (higher the similarity, higher the creativity).

\subsubsection{Critical Examination}

The Creativity Implication Network represents an effective way to deal with the creativity of sets of artworks. It considers both value and novelty, and it allows for using automated techniques in the computation of similarity. On the other hand, two potential drawbacks should be highlighted. The first one is related to artworks that occupy the position of ``leaves'' in the graph: if there are no subsequent works in the graph, their creativity would only be based on novelty, and not on value. 
The second one is more subtle, and it is about time-positioning. As demonstrated by \cite{elgammal15}, moving back an artwork has the effect of increasing its creativity; however, this appears conceptually wrong. As discussed in \cite{boden03}, the time location of an artwork is fundamental in the quantification of its creativity. It may happen that, due to the surprise component of creativity, an artwork that appears too \textit{early} might not be considered as surprising because observers are not able to truly understand it; on the contrary, if it appears too \textit{late}, it might be considered as obvious and not surprising at all. In conclusion, even if this approach is able to correctly capture value and novelty, it cannot capture the concept of surprise.

\subsection{Generate-and-Test Setting}

\subsubsection{Overview}

Generate-and-test setting \cite{toivonen15} is a family of methods based on the separation of the generative process into two phases: generation and evaluation. 
First, the system generates a candidate artifact. Then, it evaluates its degree of creativity and outputs the artifact if the evaluation is passed.
For example, the authors of \cite{varshney19a} use this approach to develop a computational creativity system for generating culinary recipes and menus called Chef Watson. \cite{colton15} describes an augmentation of Painting Fool \cite{colton12b} with Digital ARtist Communicating Intention (DARCI) \cite{norton10} in a generate-and-test setting (i.e., by using The Painting Fool for generation, and DARCI for evaluation). Pseudo-Intelligent Evolutionary Real-time Recipe Engine (PIERRE) \cite{morris12} is also based on two models, one for generating recipes with a genetic algorithm, and one for evaluating them.

\subsubsection{Dimensions of Creativity Considered}
The dimensions of creativity considered depend on the specific implementation of the evaluation function.
Different evaluation functions have been designed to evaluate, for example, quality and novelty \cite{morris12,varshney19a} and art appreciation \cite{colton15}. It is worth noting that, given the generality of the approach, other evaluation functions could be designed to capture other aspects of machine creativity.

\subsubsection{Protocol for Evaluation} \label{generateandtestprotocol}

The evaluation protocol strictly depends on the specific implementation of the \textit{eval()} function. For instance, Chef Watson \cite{varshney19a} uses two measures: flavorfulness for quality, and Bayesian surprise for novelty. Flavorfulness is computed by means of a regression model, built on olfactory pleasantness considering its constituent ingredients. Bayesian surprise \cite{baldi10} is a measure of surprise in terms of the impact of a piece of data that changes a prior distribution into a posterior distribution, calculated applying Bayes' theorem. The surprise is then the distance between the posterior and prior distributions. It is worth noting that it has been demonstrated that there exists a mathematical limit in the maximization of quality and novelty when novelty is expressed in terms of Bayesian surprise \cite{varshney19b}.
On the other hand, The Painting Fool \cite{colton12b} uses DARCI \cite{norton10} in place of the evaluation function. DARCI is able to make associations between image features and descriptions of the images learned using a series of NNs as the basis for the appreciation. It has therefore been used as a sort of artificial art critic to complement The Painting Fool, allowing it to assess the validity of its own creations.
Finally, PIERRE \cite{morris12} evaluates the generated recipes again using novelty and quality. Novelty is computed by counting new combinations of ingredients used. Quality is based on two NNs that perform a regression of user ratings based on the amount of different ingredients, working at two levels of abstraction.

\subsubsection{Critical Examination}

The advantage of the generate-and-test setting is that a variety of evaluation functions can be defined. This allows, for instance, to evaluate the generative system by means of Boden's three criteria, while still considering the specific characteristics of the domain of interest. However, its applicability is not general: as we have seen in the previous section, many generative systems do not follow the proposed setting (e.g., input-based methods use the evaluation to guide the generation, thus merging the two stages).

\subsection{Atoms of EVE'}

\subsubsection{Overview}

\cite{burns06} proposes an approach to measure aesthetic experience called Atoms of EVE', which is based on a probabilistic model of the world to derive expectations and explanations. The authors state that aesthetic arises in two ways: by forming E (\textit{expectation}) while avoiding V (\textit{violation}); and by forming E' (\textit{explanation}) while resolving V.

\subsubsection{Dimensions of Creativity Considered}

Even if not explicitly considered, the three grounding concepts of Atoms of EVE' strongly intertwine with creativity. Expectation is close to value: it measures how much we are able to understand the object of interest. Violation is close to surprise: it is the unexpectedness of an object at a certain moment. Explanation is again close to value: it measures intelligibility (i.e., its usefulness). These same considerations have been expressed by \cite{burns15}, where the author uses EVE' to define a creativity measure based on feelings (i.e., surprise), computed by means of violation, and meanings (i.e., value), computed by means of explanation.

\subsubsection{Protocol for Evaluation}

Expectation is computed as the posterior probability after the occurrence of a given object: it measures how much the prior belief can help explain that object. Violation is instead computed as the unexpectedness of that object. Together with apprehension, which is the unpredictability of the next object (before seeing it), violation returns the \textit{tension}, one of the two fundamental measures of aesthetics. Finally, explanation measures how much the encountered violation can be explained by the posterior belief. Together with expectation, explanation returns a quantification of \textit{pleasure}, the other fundamental measure of aesthetics.

\subsubsection{Critical Examination}

As observed before, while such a computation for surprise is common, this is not true for value. The focus on explanation provides an interesting way to mathematically define value. However, value is not only about finding meaning but also about utility, performance, and attractiveness \cite{maher10}; this is not possible through this measure. Finally, novelty is not considered.

\subsection{Common Model of Creativity for Design}

\subsubsection{Overview}

The authors of \cite{maher12} propose a model to evaluate creativity in the specific domain of design. They consider creativity as a relative measure in a conceptual space of potential and existing designs. Designs are represented by attribute-value pairs; and novelty, value, and surprise capture different aspects of creativity in their space. A similar approach is at the basis of the regent-dependent model \cite{franca16}, according to which artifacts are described through sets of pairs with a regent (e.g., an action or an attribute) and a dependent (e.g., the specific target or value of the regent).

\subsubsection{Dimensions of Creativity Considered}

The model proposed by \cite{maher12} considers all three dimensions suggested by \cite{boden03}, i.e., novelty, value, and surprise. Novelty is considered as a matter of comparing objects in a descriptive space; it is the degree of difference. Value is related to performance, i.e., utility preferences associated with the attributes of an object. Finally, surprise is linked to violated expectations.

\subsubsection{Protocol for Evaluation}

The model is based on the analysis of the conceptual space of potential and existing designs defined by all the potential attribute-value pairs. Novelty is evaluated with respect to a description space, i.e., by considering each product as the set of its descriptive attributes. Value is considered with respect to a performance space, i.e., by considering attributes that have utility preferences associated to them. Finally, surprise is based on finding violations of patterns that are possible to anticipate in the space of both current and possible designs. The K-means clustering algorithm is used to organize known designs by means of their attributes. Then, novelty, value, and surprise measures of a new design are obtained by looking at the distance to the nearest cluster centroid.

\subsubsection{Critical Examination} \label{attributecritic}

Even if it explicitly targets the design domain, this approach is able to combine the three dimensions of creativity by Boden. Nonetheless, it is limited by the fact that artifacts have to be described through an attribute-value pair representation. In particular, a large number of features might be needed. Otherwise, we might lose aspects of the artifacts that are fundamental to correctly quantify creativity. Since it is not possible to know the fundamental features in advance, the method requires one to enumerate as many features as possible. However, the risk is to define an excessive number of non-informative attributes, making the computation of the metrics too computationally expensive. In fact, the data points become increasingly ``sparse'' as dimensionality increases; many techniques (especially clustering) are based on distance, and therefore they may suffer from the curse of dimensionality \cite{steinbach04}. Finally, as for classic machine learning techniques, there is the need to manually define and extract the chosen features from unstructured data, which is a time-consuming and potentially prone-to-error activity. A possible way to overcome the problems related to feature extraction and the curse of dimensionality might be to adopt deep learning techniques, given their effectiveness with unstructured data.

\subsection{Unexpectedness Framework}

\subsubsection{Overview}

The authors of \cite{grace14} suggest that a framework of unexpectedness (i.e., violation of an observer's expectations) can deal with novelty, surprise, and domain transformation (also called transformativity). Although they do not claim it can be a measure of creativity on its own, and that value should be added as well, they suggest it can become a vital component in computational creativity evaluation.

\subsubsection{Dimensions of Creativity Considered}

The authors of \cite{grace14} suggest that unexpectedness can be used to compute three dimensions of creativity: novelty, surprise, and transformativity. Indeed, novelty is about the possibility of violating the observer's expectations about the continuity of a domain; if the current model of belief is not applicable to the current artifact, it can be considered novel. Surprise is instead about the possibility of violating the observer's expectations about an artifact. Finally, transformativity is about the possibility of violating the observer’s expectations about the conceptual space itself (i.e., finding that the rules governing it were not accurate).

\subsubsection{Protocol for Evaluation}

The unexpectedness framework should allow one to model expectation. Notably, expectation should be linked with the socio-cultural context of the observer, since it is the observer that forms expectation, not the domain itself. In particular, an expectation is generated by a prediction about the predicted (i.e., the dependent variables of the artifact) given a condition (i.e., a relationship between the predicted property and some other property of the object) that applies within a scope (i.e., the set of possible artifacts to which the expectations apply). An observation that falls within that scope can then be measured for congruence with respect to that expectation.

\subsubsection{Critical Examination}

The unexpectedness measure appears to be able to provide researchers and practitioners with a way to derive novelty and surprise. Notably, it also captures transformativity, clarifying at the same time how \textit{simple} surprise differs from it, i.e., that surprise is related to expectations about a single artifact, while transformativity is related to expectations about the entire domain. However, it requires defining its conceptual space in terms of explicit rules, which can be violated (and in a way that allows a violation to be detected). In addition, it does not include value in the assessment of machine creativity.

\subsection{Essential Criteria of Creativity}

\subsubsection{Overview}

The metric proposed by \cite{maher10} is based on three components: novelty, value, and surprise. It relies on the idea that a creativity metric has to be independent not only from the domain but also from the producer.

\subsubsection{Dimensions of Creativity Considered}

The criteria of creativity defined by \cite{maher10} cover exactly Boden's three criteria. In particular, novelty is intended here as a measure of how different the artifact is from known artifacts in its class. Value is quantified by considering how the potentially creative artifact compares in utility, performance, or attractiveness to other artifacts in its class. Finally, unexpectedness is defined as the variation from expectation developed for the next new artifact in its class.

\subsubsection{Protocol for Evaluation}

Novelty is calculated as the distance between the artifact of interest and the other artifacts in its class. The partition into classes is obtained by means of a clustering algorithm. Surprise is calculated by considering whether or not the artifact agrees with the expected next artifact in the pattern extracted from recent artifacts. More specifically, it is calculated as the difference between the expected next artifact and the real next artifact. Such a pattern is predicted by a self-supervised neural network; predictions are refined using reinforcement learning to correct the learned trajectory in case of sequential data. Finally, value is calculated as the weighted sum of the performance variables of the artifact. The weights depend on a co-evolutionary algorithm with a fitness function that can change over time in case the current population of artifacts changes.

\subsubsection{Critical Examination}

The method considers all three of Boden's criteria; it is not linked to a specific domain, or the producer itself; it deals with the evolution of creativity, capturing its volatile nature at the same time. However, in our opinion, it is limited in terms of applicability by the fact that it requires the definition of performance variables (similarly to other approaches based on attribute-value pairs, see Section \ref{attributecritic}).
Moreover, the setting of the parameters of the clustering algorithms at the basis of this method and the definition of distances among artifacts require human fine-tuning.

\subsection{Computational Metrics for Storytelling}

\subsubsection{Overview}

For the specific case of storytelling, the authors of \cite{karampiperis14} propose a set of computational metrics to compute the evaluation of novelty, surprise, rarity, and recreational effort.

\subsubsection{Dimensions of Creativity Considered}

Novelty and surprise are evaluated according to the standard Boden's definition, while rarity is intended as the presence of rare combinations of properties and recreational effort as the difficulty in achieving a specific result.

\subsubsection{Protocol for Evaluation}

Novelty is computed as the average semantic distance between the dominant terms included in the textual representation of the story, compared to the average semantic distance of the dominant terms in all stories. Surprise is computed as the average semantic distance between the consecutive fragments of each story. Rarity is computed as the distance between the individual clusters of each term in each story and those in the story set. Finally, recreational effort is computed as the number of different clusters each story contains.

\subsubsection{Critical Examination}

Although value is not considered, the proposed metrics appear to be appropriate to evaluate novelty and surprise. Nonetheless, they suffer from two problems: they are intrinsically domain-specific and they require that all the types of clusters are defined correctly, which is very difficult to ensure in practice.

\section{Outlook and Conclusion} \label{conclusion}

In this survey, we have provided the reader with an overview of the state of the art at the intersection between creativity and machine learning. Firstly, we have introduced the concept of machine creativity, including key concepts and definitions. Secondly, we have described a variety of generative learning techniques, considering their potential applications and limitations. Finally, we have discussed several evaluation frameworks for quantifying machine creativity, highlighting their characteristics and the dimensions they are able to capture.

Even if the field of machine creativity has witnessed increasing interest and popularity in recent years, there are still several open challenges. First of all, an interesting direction is the exploration of  \textit{creativity-oriented} objective functions, to directly train models to be creative or to navigate the induced latent space to find creative solutions. Another open problem is the definition of techniques to explore or transform the space of solutions. 
A fundamental area is the definition of novel and robust evaluation techniques for both generated and real artifacts. As discussed in Section \ref{creativitymeasures}, deep learning might be used as a basis for the definition of metrics for machine creativity.
It should be noted that there is also an ongoing debate about the role of human versus machine evaluation \cite{lamb18}.
Another promising research direction concerns the machine interpretation of art \cite{achlioptas21}. Moreover, machine learning techniques might also be used to investigate psychological dimensions of creativity~\cite{agnoli19}.
There are also foundational questions related to generative deep learning and copyright \cite{franceschelli22}. For example, it is not clear if machine-generated works could be protected by Intellectual Property, and, if they are, who should be the owner of the related rights \cite{samuelson86}. In addition, other problems concerning copyright should be considered, such as if and when training over protected work is permitted \cite{sobel20}. 
Another important ongoing debate is about authorship and the human role in creative fields in the era of AI\footnote{This is one of the ethical dilemmas highlighted by UNESCO in its \href{https://unesdoc.unesco.org/ark:/48223/pf0000367823}{Preliminary study on the Ethics of Artificial Intelligence}.}.

The models and framework discussed in this work show the remarkable potential of generative learning for machine creativity. We hope that this survey will represent a valuable guide for researchers and practitioners working in this fascinating area.

\bibliographystyle{ACM-Reference-Format}
\bibliography{main}

\end{document}